\documentclass[10pt,twocolumn,letterpaper]{article}

\usepackage{cvpr}              

\usepackage{scalerel}
\usepackage{dblfloatfix}
\usepackage{algorithm}
\usepackage{algpseudocode}
\usepackage{wrapfig}
\usepackage{tikz}
\usepackage{pgfplots}
\usepackage{pgfplotstable}
\newcommand{\circnum}[1]{%
  \tikz[baseline=(C.base)]\node[draw, circle, inner sep=0.6pt,
    minimum size=1.6ex, line width=0.4pt](C){\scriptsize #1};%
}
\algnewcommand\Input{\item[\textbf{Input:}]}
\algnewcommand\Output{\item[\textbf{Output:}]}
\newcommand{\fire}{\protect\scalerel*{\includegraphics{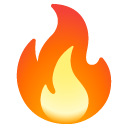}}{H}}
\newcommand{\snowflake}{\protect\scalerel*{\includegraphics{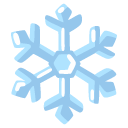}}{H}}
\newcommand\blfootnote[1]{%
  \begingroup
  \renewcommand\thefootnote{}\footnote{#1}%
  \addtocounter{footnote}{-1}%
  \endgroup
}

\usepackage{stmaryrd} 
\usepackage{amsmath}

\usepackage{algorithm}
\usepackage{algpseudocode}

\usepackage[utf8]{inputenc} 
\usepackage[T1]{fontenc}    
\usepackage{url}            
\usepackage{booktabs}       
\usepackage{amsfonts}       
\usepackage{nicefrac}       
\usepackage{microtype}      
\usepackage{xcolor}       
\usepackage{xkcdcolors}
\usepackage{times}
\usepackage{epsfig}
\usepackage{graphicx}
\usepackage{placeins}
\usepackage{multirow}
\usepackage{caption}
\usepackage[skip=5pt]{caption}
\usepackage{rotating}
\usepackage{cancel}
\usepackage{setspace}

\usepackage{bm}
\usepackage{bibunits} 

\usepackage{amssymb,amsthm}
\usepackage{amsmath}
\usepackage{graphicx}
\usepackage{wrapfig,lipsum,booktabs}

\usepackage{epsfig}
\usepackage{tikz}
\usetikzlibrary{spy}
\usepackage{algpseudocode}
\usepackage{algorithm}
\usepackage{mathrsfs}

\usepackage{nicefrac}       
\usepackage{booktabs}       

\usepackage{thmtools,thm-restate}
\usepackage{listings}
\usepackage{lstautogobble}  %
\usepackage{color}          %
\usepackage{zi4}

\usepackage{indentfirst}

\usepackage{makecell}
\usepackage{tabularx}
\usepackage{animate}
\usepackage{capt-of}

\usepackage{colortbl} 
\usepackage[table]{xcolor}

\definecolor{cvprblue}{rgb}{0.21,0.49,0.74}
\usepackage[pagebackref,breaklinks,colorlinks,allcolors=cvprblue]{hyperref}

\def\paperID{} 
\def\confName{CVPRW}
\def\confYear{2026}

\title{TeDiO: \underline{Te}mporal \underline{Di}agonal \underline{O}ptimization for Training-Free Coherent Video Diffusion}

\author{Nurislam Tursynbek$^{1*}$,\hspace{0.45cm}    Zhiqiang Lao$^{2}$,\hspace{0.45cm} Heather Yu$^{2}$,\hspace{0.45cm} Gedas Bertasius$^{1}$,\hspace{0.45cm} Marc Niethammer$^{3}$\\
UNC Chapel Hill$^1$,\hspace{0.9cm} Futurewei Technologies Inc$^2$,\hspace{0.9cm}  UCSD$^3$\\
\tt\small{nurislam@cs.unc.edu,}\hspace{0.9cm}
\tt\small{\{zlao,hyu\}@futurewei.com}
}

\begin{document}

\twocolumn[{%
\renewcommand\twocolumn[1][]{#1}%
\maketitle
\begin{center}
    \newcommand{\numColumns}{4}
    \newcommand{\columnSpacing}{0.1cm}
    \begin{tabularx}{\textwidth}{XXXX}
        \centering \small \textbf{Wan2.1} & 
        \centering \small \textbf{Wan2.1 + TeDiO (Ours)} & 
        \centering \small \textbf{Wan2.1} &
        \centering \small \textbf{Wan2.1 + TeDiO (Ours)}
    \end{tabularx}
    \begin{tabular}{
        @{}
        p{\dimexpr(\textwidth-\columnSpacing*(\numColumns-1))/\numColumns} @{\hspace{\columnSpacing}}
        p{\dimexpr(\textwidth-\columnSpacing*(\numColumns-1))/\numColumns} @{\hspace{\columnSpacing}}
        p{\dimexpr(\textwidth-\columnSpacing*(\numColumns-1))/\numColumns} @{\hspace{\columnSpacing}}
        p{\dimexpr(\textwidth-\columnSpacing*(\numColumns-1))/\numColumns} @{}
    }
        \animategraphics[loop, width=\linewidth]{32}{small_videos/man_wan/}{1}{81} &
        \animategraphics[loop, width=\linewidth]{32}{small_videos/man_tedio/}{1}{81} &
        \animategraphics[loop, width=\linewidth]{32}{small_videos/closeup_wan/}{1}{81} &
        \animategraphics[loop, width=\linewidth]{32}{small_videos/closeup_tedio/}{1}{81}
    \end{tabular}
    \begin{tabularx}{\textwidth}{XX}
        \centering \small {\fontfamily{phv}\selectfont "A man jumping on a trampoline."} & 
        \centering \small {\fontfamily{phv}\selectfont "A close-up of a runner’s legs as they sprint through a crowded city street, dodging pedestrians and street vendors"}
    \end{tabularx}

    \vspace{0.5em}
    \begin{tabularx}{\textwidth}{XXXX}
        \centering \small \textbf{CogVideoX} & 
        \centering \small \textbf{CogVideoX + TeDiO (Ours)} & 
        \centering \small \textbf{CogVideoX} &
        \centering \small \textbf{CogVideoX + TeDiO (Ours)}
    \end{tabularx}
    \renewcommand{\numColumns}{4}
    \renewcommand{\columnSpacing}{0.25em}
    \begin{tabular}{
        @{}
        p{\dimexpr(\textwidth-\columnSpacing*(\numColumns-1))/\numColumns} @{\hspace{\columnSpacing}}
        p{\dimexpr(\textwidth-\columnSpacing*(\numColumns-1))/\numColumns} @{\hspace{\columnSpacing}}
        p{\dimexpr(\textwidth-\columnSpacing*(\numColumns-1))/\numColumns} @{\hspace{\columnSpacing}}
        p{\dimexpr(\textwidth-\columnSpacing*(\numColumns-1))/\numColumns} @{}
    }
        \animategraphics[loop, width=\linewidth]{32}{small_videos/origami_cog/}{1}{81} &
        \animategraphics[loop, width=\linewidth]{32}{small_videos/origami_tedio/}{1}{81} &
        \animategraphics[loop, width=\linewidth]{32}{small_videos/panda_cog/}{1}{81} &
        \animategraphics[loop, width=\linewidth]{32}{small_videos/panda_tedio/}{2}{81}
    \end{tabular}
    \begin{tabularx}{\textwidth}{XX}
        \centering \small {\fontfamily{phv}\selectfont "Origami dancers in white paper, 3D render, on white background, studio."} & 
        \centering \small {\fontfamily{phv}\selectfont "A panda playing with a swing set."}
    \end{tabularx}
    
    
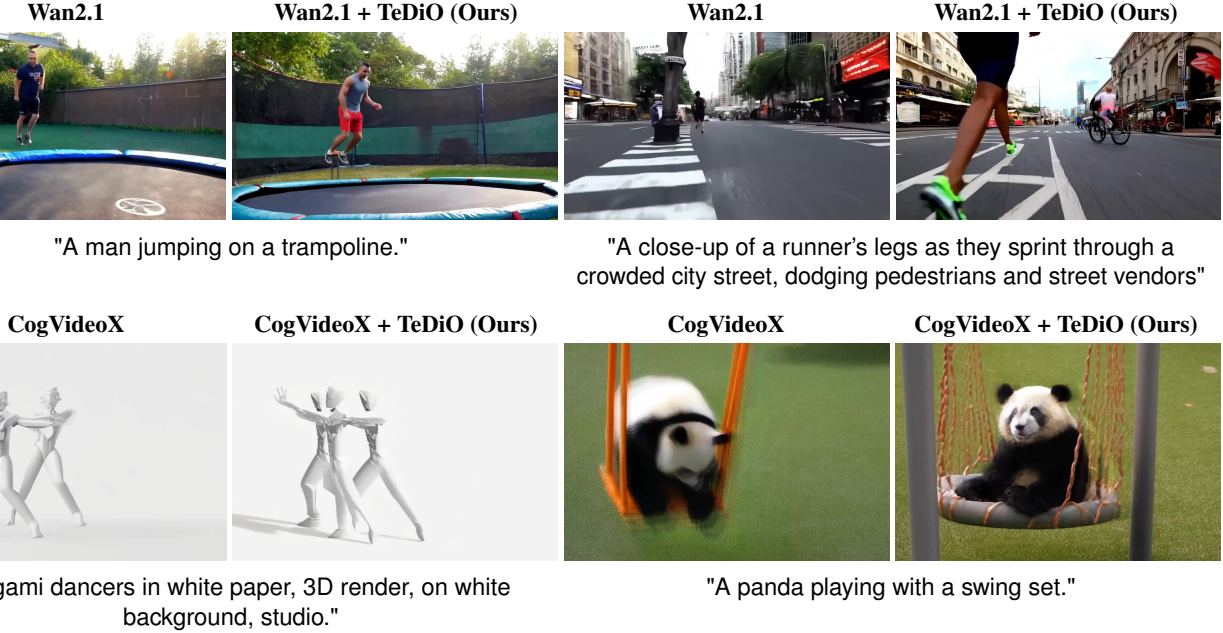
\captionof{figure}{
    \textbf{TeDiO}, our training-free inference-time method, substantially improves motion coherence in state-of-the-art text-to-video diffusion transformers. Shown are comparisons on \textbf{Wan2.1} and \textbf{CogVideoX}: without TeDiO, generations suffer from jittery inconsistent motion, merging objects, subject duplication, and physically implausible dynamics. TeDiO resolves these artifacts --- producing temporally stable, coherent videos without any finetuning, retraining, or architectural changes.\\ \textit{Best viewed in Acrobat Reader. Click to play each video clip. Note: scroll down and back up to reset a playing clip before starting the next.}}
    \label{fig:main}
\end{center}
}]

\begin{abstract}
\blfootnote{
\textsuperscript{*}Work was done during an internship at Futurewei Inc.}
Recent text-to-video diffusion transformers generate visually compelling frames, yet still struggle with temporal coherence, often producing flickering, drifting, or unstable motion. We show that these failures leave a clear imprint inside the model: incoherent videos consistently exhibit irregular, fragmented temporal diagonals in their intermediate self-attention maps, whereas stable motion corresponds to smooth, band-diagonal patterns. Building on this observation, we introduce TeDiO, a training-free, inference-time method that reinforces temporal consistency by regularizing these internal attention patterns. TeDiO estimates diagonal smoothness, identifies unstable regions, and performs lightweight latent updates that promote coherent frame-to-frame dynamics—without modifying model weights or using external motion supervision. Across multiple video diffusion models (e.g., Wan2.1, CogVideoX), TeDiO delivers markedly smoother motion while preserving per-frame visual quality, offering an efficient plug-and-play approach to improving dynamic realism in modern video generation systems.

\end{abstract} 
\section{Introduction}
\label{sec:intro}

Generative modeling has entered a new phase with the rapid progress of diffusion models. What began as a breakthrough in high-fidelity image synthesis~\cite{ho2020denoising,dhariwal2021diffusion,rombach2022high,esser2024scaling} has quickly evolved into powerful text-to-video systems~\cite{brooks2024video,polyak2024movie,wan2025,gao2025seedance,KlingAI,veo3_deepmind_2025} capable of producing visually rich and semantically aligned dynamic scenes. These models are already enabling applications in digital content creation~\cite{veo3_deepmind_2025}, educational visualization~\cite{bi2025cd}, and immersive simulation~\cite{zhou2025holotime}. As visual fidelity improves, a key bottleneck is emerging: today’s video diffusion transformers still struggle to generate motion that is stable, coherent, and perceptually natural.

Despite producing high-quality individual frames, current text-to-video models frequently exhibit \emph{temporal incoherence}~\cite{chefer2025videojam,shaulov2025flowmo,nam2025optical}: textures flicker, objects drift or deform inconsistently, and global motion lacks continuity. This instability breaks the illusion of realism and limits practical deployment in animation, robotics, simulation, and other settings where smooth temporal dynamics are essential. Prior approaches address this by modifying architectures~\cite{tulyakov2018mocogan,jin2024video} or training models with explicit motion-consistency supervision~\cite{chefer2025videojam}, but these strategies require large-scale video datasets and costly retraining - an increasing burden as model sizes continue to grow. 

An appealing alternative is to improve temporal coherence \emph{during inference}, without changing model parameters. Recent test-time optimization approaches adapt latents or prompts to stabilize motion~\cite{shaulov2025flowmo,nam2025optical}. However, these methods depend on auxiliary components such as optical flow networks~\cite{nam2025optical}, multi-GPU compute~\cite{shaulov2025flowmo}, or complex iterative pipelines, reducing practicality and generality. This motivates a more fundamental question:
\emph{Can we efficiently improve temporal coherence using only the diffusion model itself - its own internal signals - without supervision, retraining, or external modules?} Answering this would unlock a new class of inference-time techniques that seamlessly integrate with any diffusion transformer, regardless of architecture or scale.

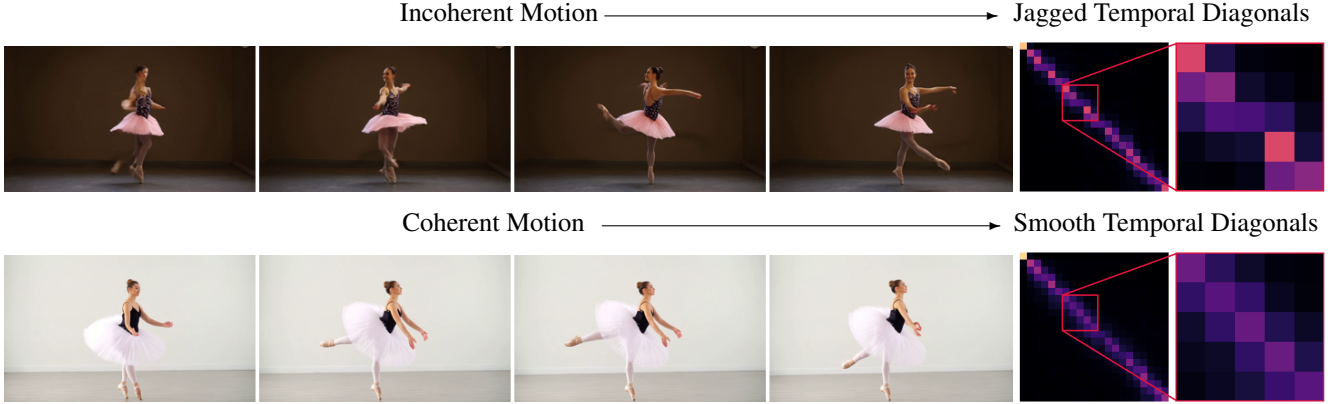
\begin{figure*}[t]
\centering
\begin{minipage}{\linewidth}
\begin{picture}(130,160)
\put(0,6){\includegraphics[width=0.77\linewidth]{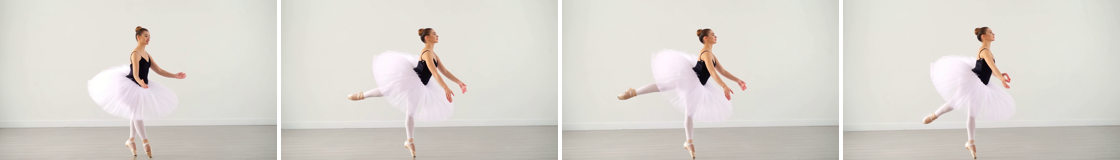}}
\put(380,0){\includegraphics[trim={0cm 0cm 0 0cm},clip,width=0.24\linewidth]{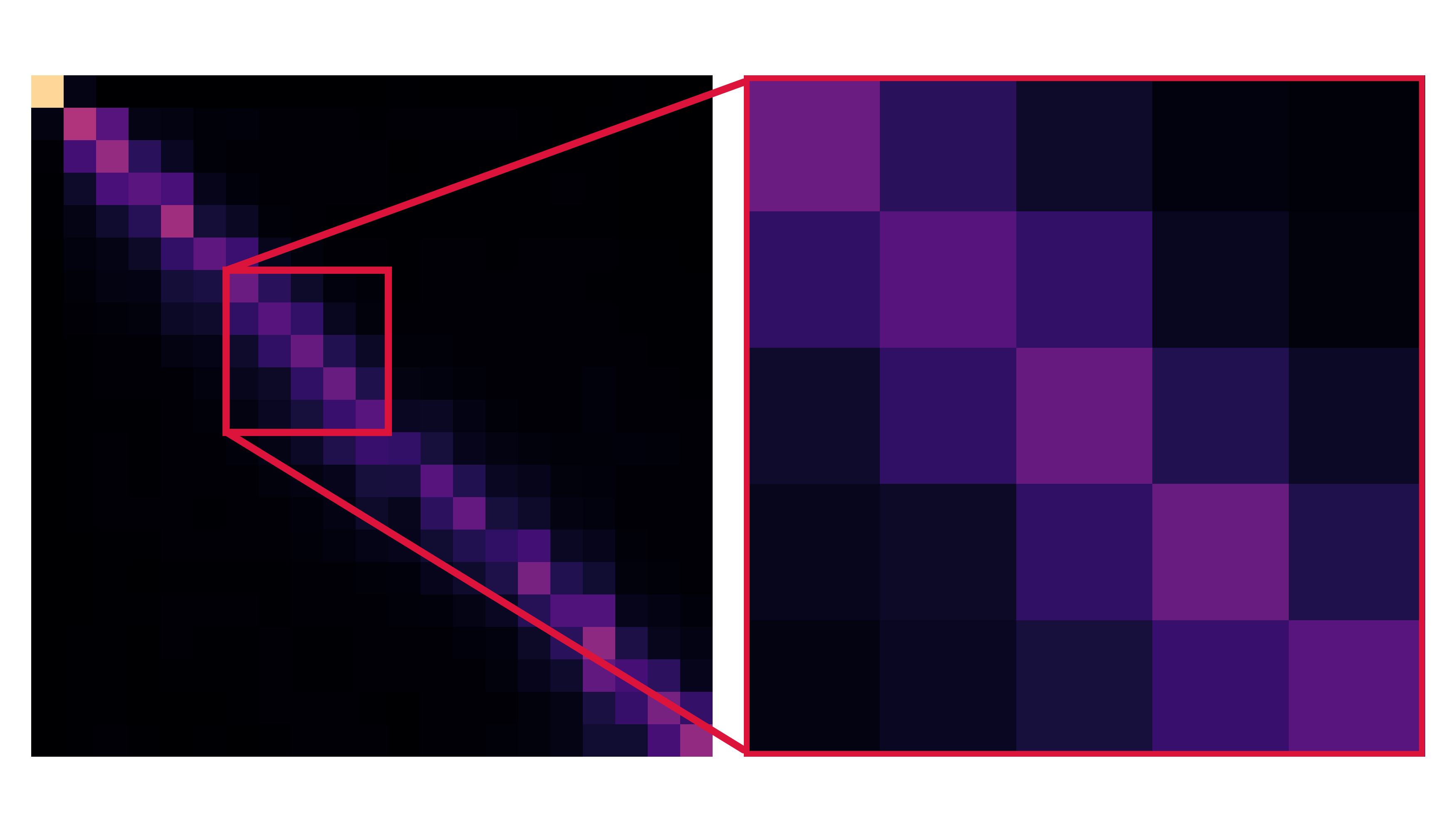}}
\put(0,85){\includegraphics[width=0.77\linewidth]{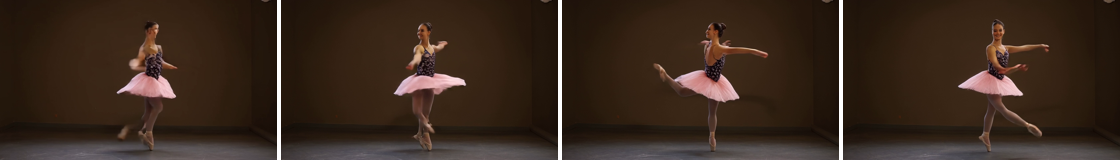}}
\put(380,79){\includegraphics[trim={0cm 0cm 0 0cm},clip,width=0.24\linewidth]{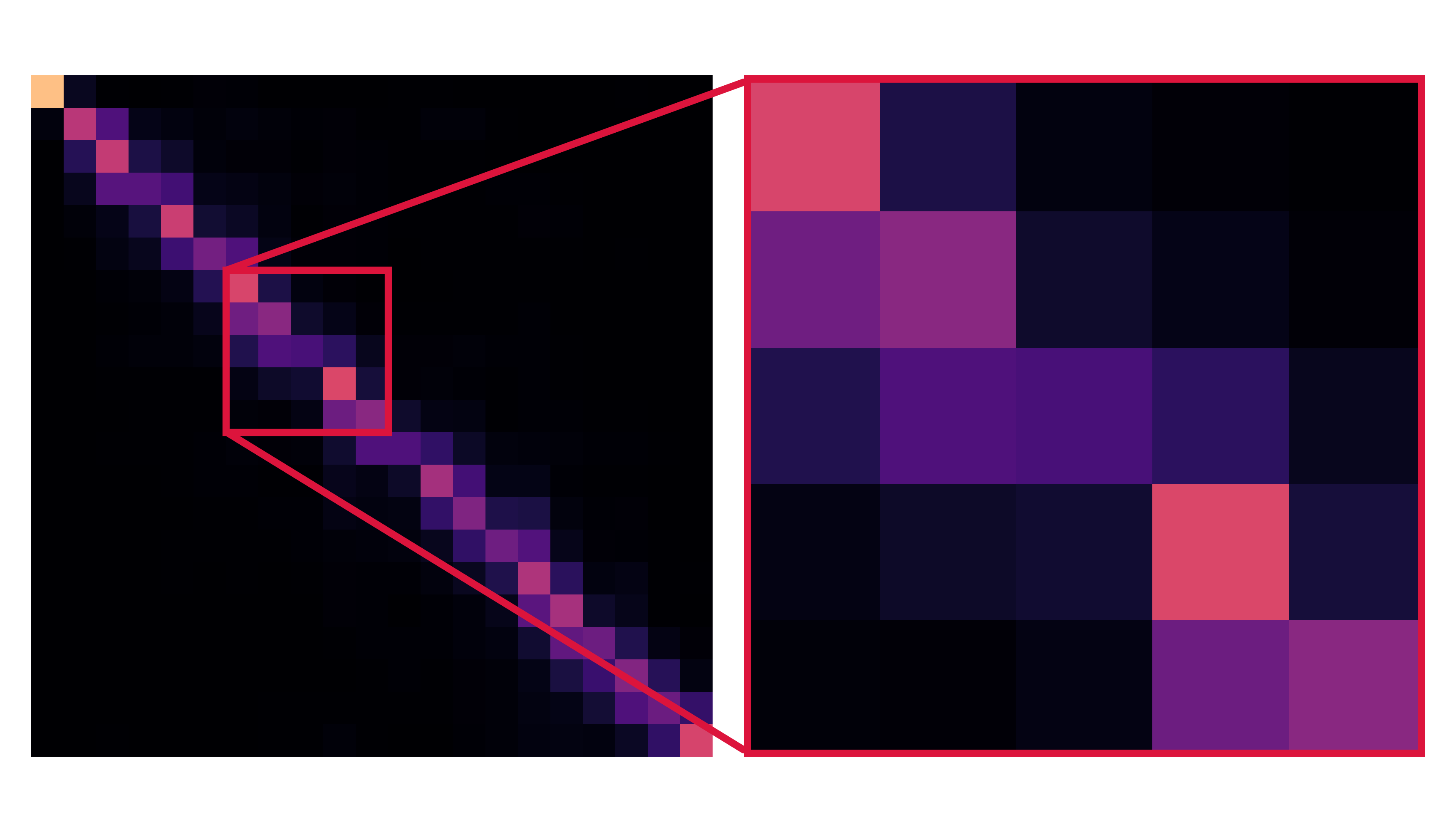}}
\put(150,70){Coherent Motion}
\put(380,70){Smooth Temporal Diagonals}
\put(149,149){Incoherent Motion}
\put(380,149){Jagged Temporal Diagonals}
\put(226,151){\vector(1,0){148}}
\put(225,72){\vector(1,0){150}}
\end{picture}
\end{minipage}
\caption{\textbf{Temporal attention maps reflects motion coherence.}
\emph{Incoherent} videos often exhibit flickering or abrupt transitions (top row), whereas \emph{coherent} videos show smooth and stable dynamics (bottom row). Their corresponding temporal attention maps reveal a structural difference: incoherent motion produces irregular, fragmented diagonals, while coherent motion shows clean, band-diagonal patterns that indicate stable frame-to-frame alignment. These internal attention structures provide an interpretable internal signal of motion stability.}
\label{fig:temporal_inconsistency}
\end{figure*}

Our starting point is a simple but crucial observation: temporal coherence leaves a clear, measurable footprint inside a diffusion transformer’s temporal self-attention. As shown in Figure~\ref{fig:temporal_inconsistency}, incoherent motion results in \emph{irregular, fragmented temporal diagonals}, reflecting unstable alignment between consecutive frames. In contrast, smooth and stable videos produce \emph{clean, band-diagonal structures} that encode consistent temporal dependencies. Coherent video generations consistently exhibit lower \emph{diagonal variability} across different text-to-video diffusion models and across multiple transformer blocks, revealing a robust and architecture-agnostic quantitative marker of motion stability. This provides a direct and interpretable link between the structure of temporal attention and the quality of generated motion - suggesting that motion stability can be diagnosed and improved \emph{from within the generation process itself}. 

Building on this insight, we propose \textbf{TeDiO} (\emph{Temporal Diagonal Optimization}), a lightweight, training-free method that enhances motion coherence by \emph{directly optimizing temporal diagonals in self-attention}. TeDiO operates only during the earliest diffusion steps - where global motion trajectories are shaped - and identifies spatial patches whose temporal attention maps exhibit irregular or noisy diagonal structure. For these patches, TeDiO applies a small latent update that promotes smooth, band-diagonal temporal attention, effectively reinforcing stable frame-to-frame alignment. The method requires no retraining, external supervision (e.g., optical flow), or multi-GPU compute, making it fully plug-and-play for existing video diffusion transformers. By intervening precisely at the internal locations where instability originates, TeDiO improves global motion while preserving the model’s inherent per-frame image quality.

TeDiO produces consistent and substantial improvements in temporal stability across modern state-of-the-art video diffusion models, including Wan2.1-1.3B~\cite{wan2025} and CogVideoX-5B~\cite{yang2024cogvideox}, with negligible computational overhead. Beyond its practical performance, TeDiO illustrates a broader paradigm: rather than scaling model size or training data, significant gains in dynamic realism can be achieved by exploiting the structure already encoded inside diffusion transformers. Our findings complement recent work on test-time latent optimization by demonstrating that temporal self-attention structures itself provide a reliable and actionable signal for improving motion coherence

In summary, this work makes three \textbf{key contributions}:
\begin{itemize}
\item We uncover a clear and quantifiable link between temporal attention diagonals and motion stability, offering a new perspective for analyzing coherence in video diffusion transformers.
\item We introduce TeDiO, a training-free inference-time approach that regularizes diagonal variability in temporal attention and optimize input latent to enhance motion coherence - without modifying model weights or relying on external supervision.
\item We provide comprehensive empirical validation across multiple diffusion backbones and benchmarks, demonstrating consistent improvements in temporal smoothness, perceptual stability, and overall video realism.
\end{itemize}

\section{Related Work}
\label{related_work}

\textbf{Diffusion models for text-to-video generation.}
Diffusion models have rapidly become the dominant framework for generative modeling, achieving remarkable fidelity in image synthesis~\cite{dhariwal2021diffusion,peebles2023scalable,rombach2022high,esser2024scaling}. Recent advances extend this paradigm to video, enabling text-conditioned generation of realistic and temporally rich scenes~\cite{ho2022imagen,singer2022make,villegas2022phenaki,blattmann2023stable,yang2024cogvideox,wan2025,gao2025seedance,KlingAI,kong2024hunyuanvideo}. These models can render visually striking clips that align closely with textual prompts, marking a major step forward in controllable video synthesis. State-of-the-art open-source text-to-video diffusion models \cite{wan2025,yang2024cogvideox} are based on Diffusion Transformer (DiT) architecture \cite{peebles2023scalable}, which serve as basis for our method. 

\textbf{Temporal consistency in video diffusion transformers} remains a persistent challenge, despite producing high-quality individual frames. Objects drift, textures flicker, and motion coherence often breaks down, undermining perceptual realism. A growing body of work seeks to improve temporal coherence in generative video models~\cite{chefer2025videojam,nam2025optical,shaulov2025flowmo,brooks2024video,kang2024far}. Training-time approaches typically impose motion-aware regularization or auxiliary supervision. For instance, VideoJam~\cite{chefer2025videojam} enforces smooth dynamics via explicit motion objectives, while other methods leverage feedback from large language models to maintain frame-to-frame consistency~\cite{zhang2025think,wu2024boosting}. Although effective, these strategies require retraining diffusion backbones, making them computationally expensive and less generalizable to existing pretrained models. Alternative solutions guide generation with external motion signals or optical flow priors~\cite{geng2025motion,ma2024trailblazer,nam2025optical,liu2024physgen,cong2023flatten}, improving stability but at the cost of additional models or discriminators. Architectural modifications—such as motion-specific modules or spatio-temporal fusion layers~\cite{tulyakov2018mocogan,villegas2022phenaki,jin2024video,he2022latent} - also enhance coherence, yet they are often tightly coupled to specific pipelines and cannot be easily transferred to powerful pretrained backbones.

\textbf{Test-time optimization} methods recently have emerged as a practical alternative for improving generative outputs without modifying model weights. This paradigm has been applied to image generation~\cite{chefer2023attend,bao2024separate,ma2025inference,bao2024separate}, where proxy losses guide sampling process during inference by optimizing input to enhance fidelity or alignment. In video generation, several works adapt latents \emph{during} denoising to improve temporal stability. MotionPrompt \cite{nam2025optical} uses external models , such as optical flow estimators and additional discriminators to guide the prompt embedding for better temporal coherence. FlowMo \cite{shaulov2025flowmo} optimizes input latent, based on patch-wise temporal variance of the output latent. This method backpropagates gradients through the entire network, resulting in $2.39\times$ longer sampling and requirement of at least 2 GPUs.  In contrast our approach improves coherence in generated videos with no external supervision and negligible computational overhead over baselines.

\textbf{Attention control in diffusion models.} A related line of image diffusion works manipulates attention maps  directly during inference~\cite{hertz2022prompt,cao2023masactrl,alaluf2024cross}, revealing that attention structure encodes rich spatial and semantic information that can be exploited post-hoc. In video generation, attention control was used for motion transfer \cite{ling2024motionclone,pondaven2025video} or editing the style or subject in the video \cite{bai2025uniedit,liu2024video,ku2024anyv2v}.  Our approach manipulates the temporal diagonals in video diffusion transformers to enhance coherence of generated video, addressing one potential source of motion inconsistency.

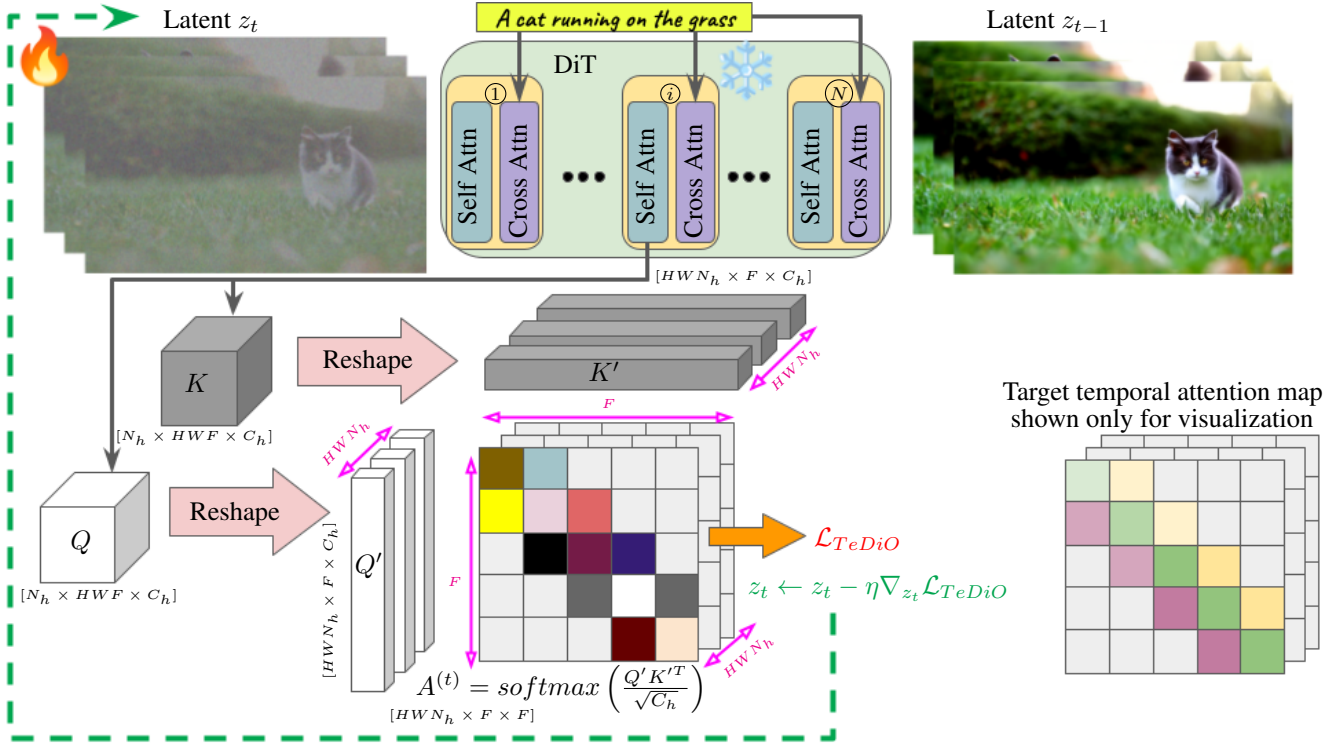
\begin{figure*}[t]
\centering
\begin{minipage}{\linewidth}
\begin{picture}(200,282)
\put(0,0){\includegraphics[width=\linewidth]{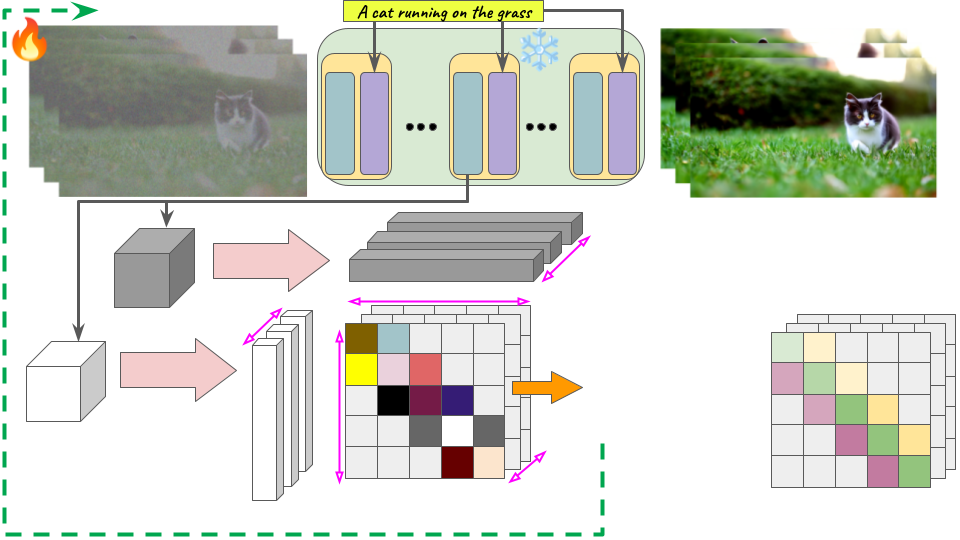}}
\put(60,270){Latent $z_t$}
\put(370,270){Latent $z_{t-1}$}
\put(181,243){\circnum{$1$}}
\put(247,243){\circnum{$i$}}
\put(309,243){\circnum{$N$}}
\put(207,253){DiT}
\put(171, 196){\rotatebox{90}{Self Attn}}
\put(190, 191){\rotatebox{90}{Cross Attn}}
\put(238, 196){\rotatebox{90}{Self Attn}}
\put(256, 191){\rotatebox{90}{Cross Attn}}
\put(300, 196){\rotatebox{90}{Self Attn}}
\put(318, 191){\rotatebox{90}{Cross Attn}}
\put(155,19){$A^{(t)}=softmax\left(\frac{Q'K'^T}{\sqrt{C_h}}\right)$}
\put(145,10){\tiny{$[HW N_h\times F\times F]$}}
\put(25,74){$Q$}
\put(132,65){$Q'$}
\put(6,55){\tiny{$[N_h\times HWF\times C_h]$}}
\put(68,133){$K$}
\put(220,137){$K'$}
\put(42,115){\tiny{$[N_h\times HWF\times C_h]$}}
\put(245,175){\tiny{$[HWN_h\times F\times C_h]$}}
\put(120,25){\rotatebox{90}{\tiny{$[HWN_h\times F\times C_h]$}}}
\put(225,127){\textcolor{Magenta}{\tiny{$F$}}}
\put(167,60){\textcolor{Magenta}{\tiny{$F$}}}
\put(119,105){\textcolor{Magenta}{\rotatebox{47}{\tiny{$HWN_h$}}}}
\put(288,133){\textcolor{Magenta}{\rotatebox{45}{\tiny{$HWN_h$}}}}
\put(270,25){\textcolor{Magenta}{\rotatebox{42}{\tiny{$HWN_h$}}}}
\put(120,142){Reshape}
\put(70,85){Reshape}
\put(305,75){\textcolor{red}{$\mathcal{L}_{TeDiO}$}}
\put(280,57){\textcolor{Green}{$z_t\leftarrow z_t - \eta\nabla_{z_t}\mathcal{L}_{TeDiO}$}}
\put(376, 130){Target temporal attention map}
\put(379, 120){shown only for visualization}
\end{picture}
\end{minipage}
\caption{\textbf{Overview of TeDiO.}
During the sampling timestep $t$, the queries ($Q$) and the keys ($K$) from block $i$ of frozen (\snowflake{}) Diffusion Transformer (DiT) are reshaped to isolate temporal interactions. A lightweight inference-time optimization adjusts the latent $z_t$ (\fire{}) using the $\mathcal{L}_{\text{TeDiO}}$ loss, encouraging smooth, band-diagonal patterns in temporal attention—indicative of stable frame-to-frame dependencies.
This process requires no retraining or auxiliary supervision and serves purely as a plug-and-play enhancement for temporal coherence.}
\label{pipeline}
\end{figure*}

\section{Method}

\subsection{Preliminaries}

Following the success of latent diffusion models for image generation~\cite{rombach2022high}, 
state-of-the-art video diffusion models~\cite{wan2025,yang2024cogvideox} 
also operate in the latent space of a pre-trained Variational Autoencoder (VAE). 
The VAE consists of an encoder $\mathcal{E}$ that maps an input video $x$ 
to a latent representation $z = \mathcal{E}(x)$ with reduced spatial and temporal dimensions, 
and a decoder $\mathcal{D}$ that approximately reconstructs the video, 
$x \approx \mathcal{D}(\mathcal{E}(x))$.

A Diffusion Transformer (DiT)~\cite{peebles2023scalable} is a neural network parameterized by $\theta$ 
that progressively transforms a Gaussian noise sample $z_T \sim \mathcal{N}(0, I)$ 
into a realistic latent $z_0$ over $T$ timesteps. 
Depending on the training formulation, DiTs are trained using either a
\emph{denoising diffusion objective} or a 
\emph{flow matching objective}.

In the denoising formulation~\cite{ho2020denoising}, adopted by CogVideoX~\cite{yang2024cogvideox}, at timestep $t$, the clean latent $\mathcal{E}(x)$ is corrupted with noise $\epsilon$ and noise-schedule coefficients $(\alpha_t, \sigma_t)$ as $z_t=\alpha_t \mathcal{E}(x) + \sigma_t \epsilon$. The model predicts the added noise $\epsilon$ and is trained with the reconstruction loss:
\begin{equation}
    \mathcal{L}_{\text{DDPM}} 
    = \| \epsilon - \epsilon_\theta(z_t, t) \|_2^2.
\end{equation}

In the flow matching formulation~\cite{lipman2022flow}, used in Wan2.1~\cite{wan2025}, the input $z_t = (1 - t/T)\mathcal{E}(x) + (t/T) z_T$ is an interpolation between Gaussian noise $z_T\sim \mathcal{N}(0, I)$ and clean input $z_0=\mathcal{E}(x)$. The model learns a velocity field $u_\theta$ that transports the noise toward the data distribution:
\begin{equation}
    \mathcal{L}_{\text{FM}} 
    = \| (z_T - z_0) 
    - u_\theta\big(z_t, t) \|_2^2.
\end{equation}
Both formulations achieve competitive generation quality and can serve as the backbone for our method. These models use a Transformer~\cite{vaswani2017attention,peebles2023scalable} consisting of $N$ blocks.

In multi-head self-attention of the block $i \in \llbracket 1, N\rrbracket$, the $Q$ and $K$ are linear projections of the input, where $Q, K \in \mathbb{R}^{N_h \times HWF \times C_h}$. $H,W,F$ denotes the spatial height, width, and number of frames of the latent, and $N_h, C_h$ are the number of attention heads and the channel dimension per head.  The original 3D attention map $A$ computes the correlation among all spatio-temporal patches:
\begin{equation}
    A = softmax\left( \frac{Q K^\top}{\sqrt{C_h}} \right)
    \in \mathbb{R}^{N_h \times HWF \times HWF}.
\end{equation}

This spatio-temporal attention structure plays a crucial role in modeling  cross-frame dependencies and, as demonstrated in the next section, is directly tied to the \emph{temporal coherence} of the generated video.

\subsection{Temporal Diagonal Optimization (TeDiO)}
\label{sec:tedio}

Video diffusion transformers use full 3D self-attention, mixing information across all spatial locations and all frames simultaneously. This global receptive field enables long-range reasoning, but it also means the model has no inductive bias encouraging smooth temporal evolution. As a consequence, attention may shift unpredictably between frames, leading to flickering or unstable motion even when the underlying scene is locally consistent.

To analyze this behavior without changing the model’s computations, we introduce a lightweight reparameterization of the attention inputs. Instead of treating the flattened spatio-temporal tokens as a single $HWF$ axis, we group all spatial positions and attention heads together while keeping the temporal dimension explicit. Importantly, this is only a reshaping of $Q$ and $K$ (into $Q', K' \in \mathbb{R}^{HWN_h \times F \times C_h}$) for diagnostic purposes - the model’s forward pass and attention computation remains unchanged:


We then compute a \emph{\textbf{temporal attention map}} for each combined spatial-head index $p\in\llbracket 1,(HWN_h)\rrbracket$: 
\begin{equation} A^{(t)}_{p} = softmax\left(\frac{Q'_{p} K'^{\top}_{p}}{\sqrt{C_h}}\right), \quad A^{(t)}_{p} \in \mathbb{R}^{F \times F}, 
\end{equation} 
which quantifies how frame $ f$ attends to other frames within the same spatial region and attention head. Collectively, these per-patch maps form a full temporal attention tensor: \begin{equation} 
A^{(t)} \in \mathbb{R}^{HWN_h \times F \times F}. 
\end{equation}

Our key observation is that coherent motion consistently produces smooth, band-diagonal patterns in $A^{(t)}$. The main diagonal reflects self-consistency across frames, while the first sub- and super-diagonals capture dependencies between consecutive frames (i.e., motion continuity). In contrast, incoherent motion produces fragmented diagonal patterns (see Figure~\ref{fig:temporal_inconsistency}), signaling abrupt representational shifts.

For each temporal attention map $A^{(t)}_{p}$, we define its diagonal band at offset $b$ as $d^{(b)}_{p}$ such that its $i$-th element:
\begin{equation}
    (d_p^{(b)})_i = (A_p^{(t)})_{i,i+b},\quad i\in\llbracket 1, F-|b|\rrbracket, \quad b \in \{-1, 0, +1\}.
\end{equation}
Here $b=0$ corresponds to the main diagonal, and $b=\pm1$ correspond to its immediate neighbors. While we empirically focus on $\{-1,0,+1\}$, as the main attention mass lies in these diagonals, the formulation naturally extends to wider temporal neighborhoods beyond first super- and sub-diagonals, i.e. $|b|\ge 2$, providing a controllable trade-off between local smoothness and global rigidity.

We quantify the regularity of each patch-head $p$ using a diagonal variability score:
\begin{equation}
\label{diag_var}
    \mathcal{S}_{(p)}
    = \sum_{b \in \{-1, 0, +1\}}
      \sum_{f=1}^{F - |b| - 1}
      \big((d^{(b)}_{p})_{f+1} - (d^{(b)}_{p})_{f}\big)^2.
\end{equation}
This score measures local fluctuations along each diagonal band: large values indicate abrupt attention shifts (i.e., “jerky” temporal behavior), while low values correspond to stable motion. $\mathcal{S}_{(p)}$ thus provides a self-diagnostic indicator of temporal inconsistency directly from the model’s internal activations. Note that this score essentially sums over gradients along a diagonal band which amounts to a smoothness loss along the diagonal band, as in the Dirichlet energy related to Gaussian smoothing.

\begin{algorithm}[t]
\caption{TeDiO diffusion sampling}
\label{alg1}
\begin{algorithmic}[1]
\Input Text prompt $\mathcal{P}$, optimization steps $\{\tau_1, \ldots, \tau_\ell\}$, and a pretrained diffusion model.
\Output Coherent video $z_0$
\State Sample $z_T \sim \mathcal{N}(0, \mathbf{I})$
\For{$t = T$ to $1$}
    \If{$t \in \{\tau_1, \ldots, \tau_\ell\}$}
        \For{$i = 1$ to $n_{iters}$}
            \State $z_t \leftarrow z_t - \eta \nabla_{z_t} \mathcal{L}_{\text{TeDiO}}$
        \EndFor
    \EndIf
    \State $z_{t-1} = \text{DiffusionStep}(z_t, \mathcal{P})$
\EndFor
\State \Return $z_0$
\end{algorithmic}
\end{algorithm}

Temporal incoherence is usually concentrated in a small subset of spatial-head entries. To focus optimization where it matters, we select the top-$k$ entries with the highest diagonal variability scores:
\begin{equation}
\label{topk}
    \mathcal{L}_{\text{TeDiO}} = \frac{1}{k}\sum_{p\in TopK(\mathcal{S}_{(p)})}\big(\mathcal{S}_{(p)}\big).
\end{equation}
This selective design makes TeDiO efficient and interpretable: it focuses computational effort where temporal noise is highest, rather than enforcing uniform regularization across the entire video.

TeDiO is applied purely at inference time as a lightweight latent optimization procedure during early diffusion timesteps. For a given timestep $t$, the latent $z_t$ is refined through a few gradient steps:
\begin{equation}
    z_t \leftarrow z_t - \eta \nabla_{z_t}
    \mathcal{L}_{TeDiO}.
\end{equation}
where $\eta$ is a small learning rate and $n_{iters}=3$ by default. This update is performed for the first $\ell$ timesteps ${\tau_1,\dots,\tau_\ell}$ out of $T$ total diffusion steps. Early-stage refinement is crucial: temporal coherence seeded in the early denoising stages propagates forward, amplifying stability throughout the generation process. This procedure is fully model-agnostic and incurs negligible additional cost. A schematic overview of TeDiO is provided in Figure~\ref{pipeline}, and its integration into diffusion sampling is summarized in Algorithm~\ref{alg1}.

\begin{table*}[t]
    \centering
    \caption{\textbf{VBench prompts evaluation} on temporal consistency and motion coherence metrics. TeDiO improves temporal coherence (smoothness and flickering) while maintaining perceptual consistency.}
    \begin{tabular}{l|cccccc}
    \toprule
    & \begin{tabular}{@{}c@{}}Motion \\ Smoothness \end{tabular} $\uparrow$ & \begin{tabular}{@{}c@{}}Dynamic \\ Degree \end{tabular} $\uparrow$& \begin{tabular}{@{}c@{}}Subject \\ Consistency \end{tabular} $\uparrow$ & \begin{tabular}{@{}c@{}}Background \\ Consistency \end{tabular}  $\uparrow$ & \begin{tabular}{@{}c@{}}Temporal \\ Flickering \end{tabular}  $\uparrow$ & \begin{tabular}{@{}c@{}}Overall \\ Consistency \end{tabular} $\uparrow$ \\
         \midrule
         Wan2.1~\cite{wan2025}& 98.32 & \textbf{56.39} & 94.85 & 98.15 & 99.38 & \textbf{23.07} \\
          +\textbf{TeDiO (ours)} & \textbf{98.78} & 29.72 & \textbf{97.05} & \textbf{98.71} & \textbf{99.47} & 22.81 \\
          \midrule
          CogVideoX~\cite{yang2024cogvideox}   & 97.81 & \textbf{81.39} & 93.84 & 94.53 & 97.99 & \textbf{23.79} \\
          +\textbf{TeDiO (ours)} & \textbf{98.26} & 68.61 & \textbf{95.37} & \textbf{96.17} & \textbf{98.87} & 23.29\\
         \bottomrule
    \end{tabular}
    \label{tab:vbench}
\end{table*}

\section{Experiments}

\subsection{Setup}

We evaluate TeDiO on two state-of-the-art open-source text-to-video DiTs representing both training paradigms:
\begin{itemize}[leftmargin=*, topsep=0cm, itemsep=0cm, partopsep=0cm, parsep=0cm] 
\item \textbf{Wan2.1-1.3B}~\cite{wan2025} trained with the flow matching objective~\cite{lipman2022flow} and evaluated at a resolution of $480\times832$
\item \textbf{CogVideoX-5B}~\cite{yang2024cogvideox} trained with the denoising objective~\cite{ho2020denoising}  and evaluated at a resolution of $480\times720$
\end{itemize}
Both models generate $81$ frames at $16$~fps (5-second clips) with the default 50-step DDPM sampler. TeDiO is applied only during the first few diffusion steps. All experiments are done on a single NVIDIA H100 GPU (80GB) using the official \texttt{PyTorch} and \texttt{diffusers}~\cite{von-platen-etal-2022-diffusers} implementations. 

\subsection{Results}
Following prior work on motion coherence~\cite{chefer2025videojam, shaulov2025flowmo}, we evaluate on two complementary prompt sets:

\begin{itemize}[leftmargin=*, topsep=0cm, itemsep=0cm, partopsep=0cm, parsep=0cm] 
\item \textbf{VBench}~\cite{huang2024vbench}: standard  benchmark assessing video generation quality across perceptual and temporal axes, containing many static or weakly dynamic prompts. This makes it a valuable testbed for assessing whether a method preserves visual quality while improving stability.
\item \textbf{VideoJAM-Bench}~\cite{chefer2025videojam}: 128 deliberately challenging prompts designed to expose temporal instability, fast motion, and complex multi-object interactions, making it a much more demanding test of temporal coherence..
\end{itemize}

\vspace{0.1cm}
\noindent We focus on the following VBench metrics \cite{huang2024vbench} that directly capture motion coherence and temporal consistency:

\begin{itemize}[leftmargin=*, topsep=0cm, itemsep=0cm, partopsep=0cm, parsep=0cm] 
\item \textbf{Motion smoothness}: Even-numbered frames interpolation error from odd-numbered frames with AMT~\cite{li2023amt}. 
\item \textbf{Dynamic degree}: top 5\% of pixels with highest magnitude of optical flow strength with RAFT \cite{teed2020raft}.
\item \textbf{Subject consistency}: DINO~\cite{caron2021emerging} similarity between consecutive frames and between the first frame and all others.
\item \textbf{Background consistency}: CLIP~\cite{radford2021learning} similarity between consecutive frames and the first frame and all others. 
\item \textbf{Temporal Flickering}: Mean Absolute Difference between consecutive frames for static prompts. 
\item \textbf{Overall Consistency}: text alignment with ViCLIP~\cite{wang2023internvid}.
\end{itemize}

\vspace{0.1cm}

\noindent For custom prompts (such as VideoJAM-Bench~\cite{chefer2025videojam}) Temporal Flickering and Overall Consistency are unavailable. Instead we report perceptual frame quality metrics:

\begin{itemize}[leftmargin=*, topsep=0cm, itemsep=0cm, partopsep=0cm, parsep=0cm] 
\item \textbf{Imaging Quality}: Per-frame MUSIQ score~\cite{ke2021musiq}.
\item \textbf{Aesthetic Quality}: Per-frame LAION score~\cite{LAION}.
\end{itemize}

Full definitions and evaluation protocols follow the official benchmark specifications.



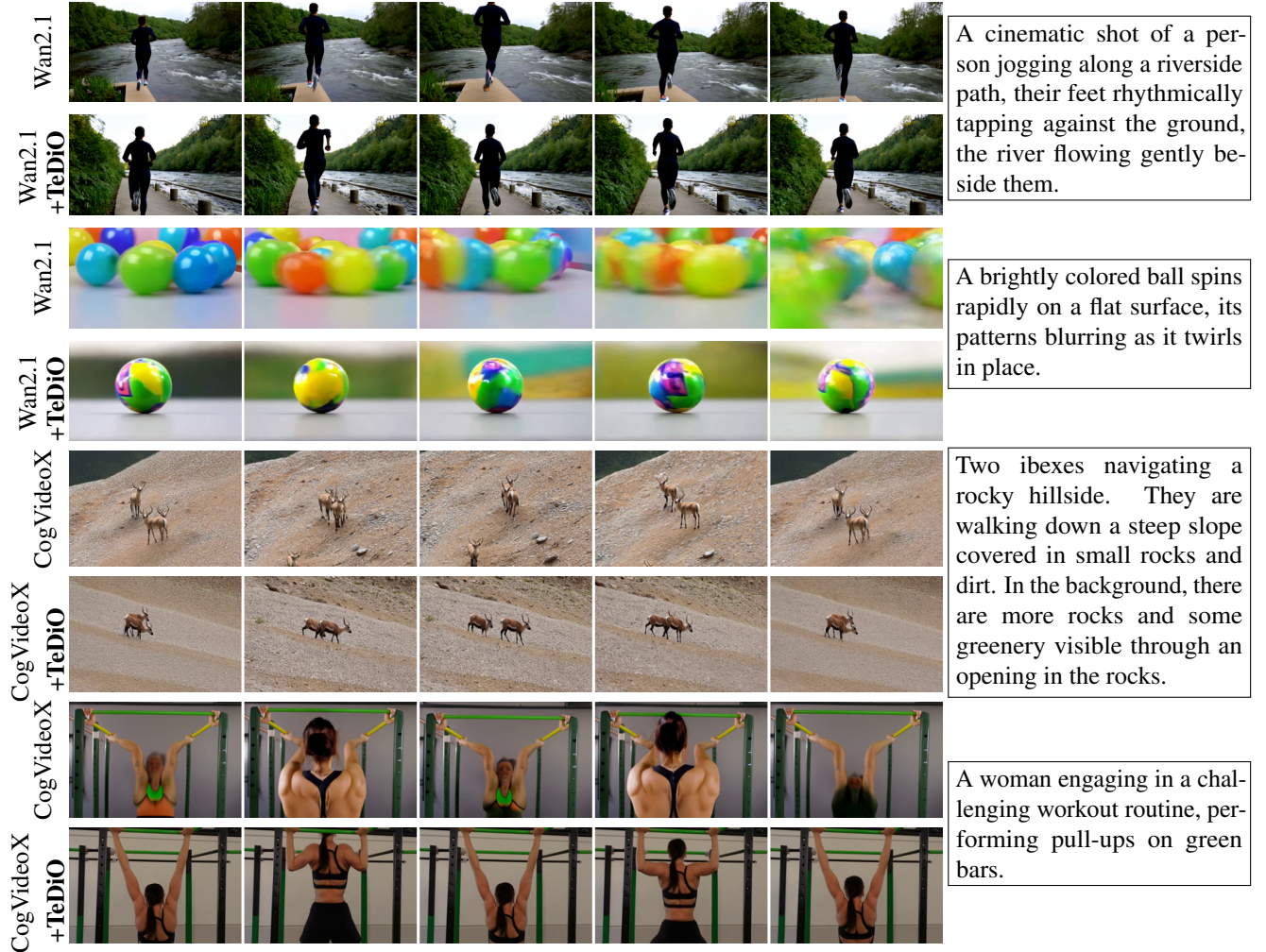
\begin{figure*}[t]
    \centering
    \begin{minipage}{\linewidth}
\begin{picture}(200,370)
\put(25,335){\includegraphics[width=0.7\linewidth]{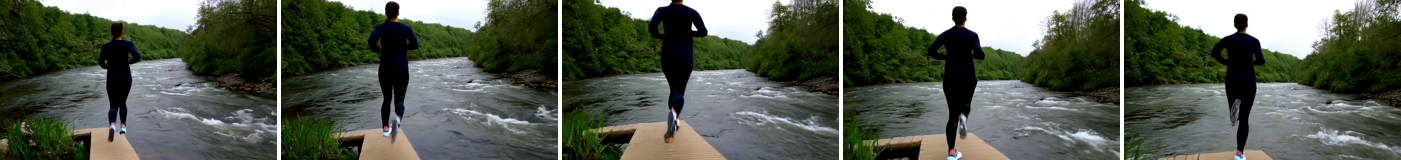}}
\put(25,290){\includegraphics[width=0.7\linewidth]{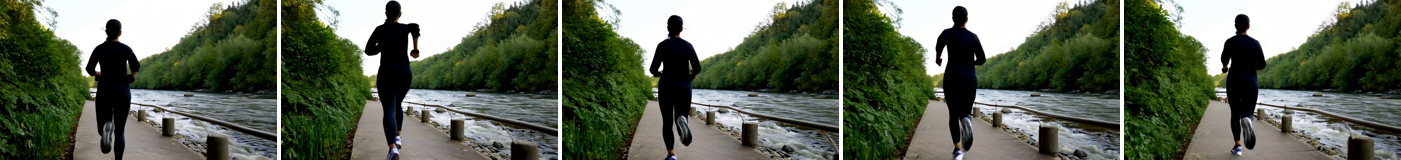}}
\put(11, 340){\rotatebox{90}{Wan2.1}}
\put(5, 294){\rotatebox{90}{Wan2.1}}
\put(16, 290){\rotatebox{90}{+\textbf{TeDiO}}}
\put(375,330){\fbox{\parbox{114pt}{
                A cinematic shot of a person jogging along a riverside path, their feet rhythmically tapping against the ground, the river flowing gently beside them. 
            }}}
\put(25,245){\includegraphics[width=0.7\linewidth]{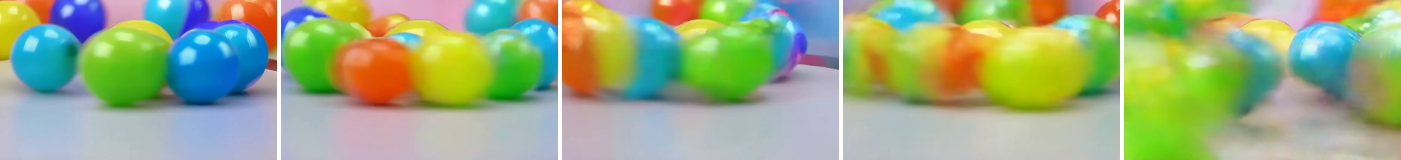}}
\put(25,200){\includegraphics[width=0.7\linewidth]{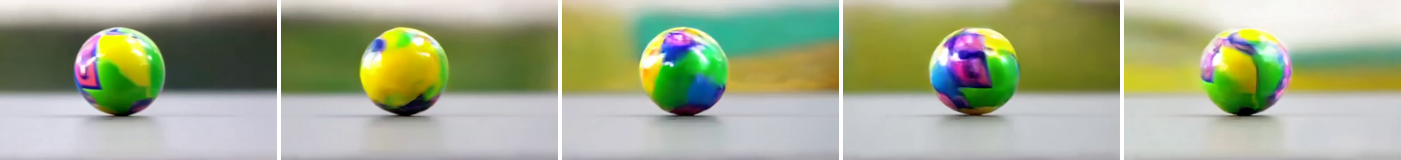}}
\put(11, 250){\rotatebox{90}{Wan2.1}}
\put(5, 204){\rotatebox{90}{Wan2.1}}
\put(16, 200){\rotatebox{90}{+\textbf{TeDiO}}}
\put(375,244){\fbox{\parbox{114pt}{
                A brightly colored ball spins rapidly on a flat surface, its patterns blurring as it twirls in place.}}}
\put(25,150){\includegraphics[width=0.7\linewidth]{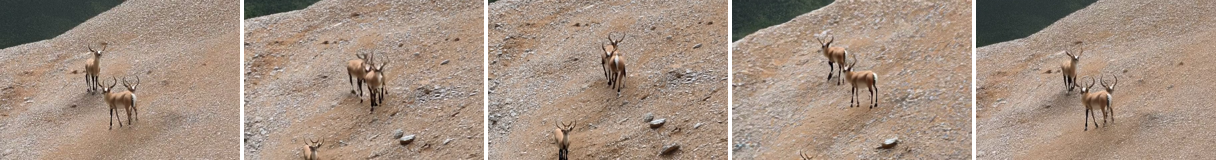}}
\put(25,100){\includegraphics[width=0.7\linewidth]{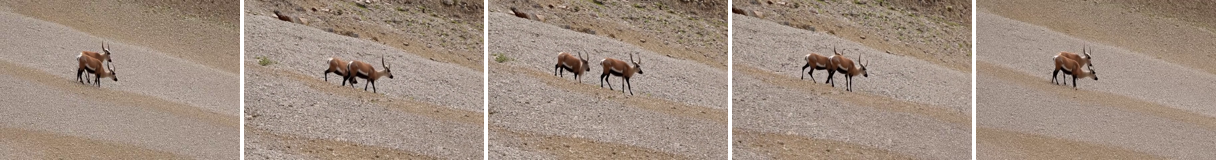}}
\put(11, 149){\rotatebox{90}{CogVideoX}}
\put(2, 96){\rotatebox{90}{CogVideoX}}
\put(16, 100){\rotatebox{90}{+\textbf{TeDiO}}}
\put(375,145){\fbox{\parbox{114pt}{
                Two ibexes navigating a rocky hillside. They are walking down a steep slope covered in small rocks and dirt. In the background, there are more rocks and some greenery visible through an opening in the rocks.
            }}}
\put(25,50){\includegraphics[width=0.7\linewidth]{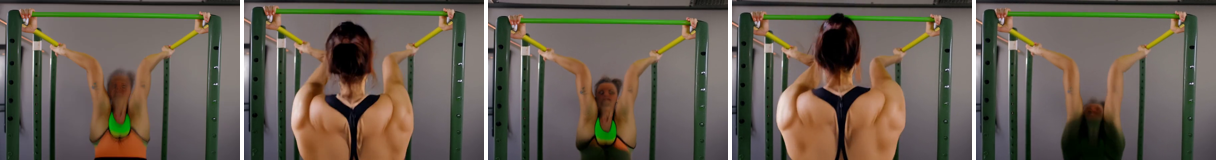}}
\put(25,0){\includegraphics[width=0.7\linewidth]{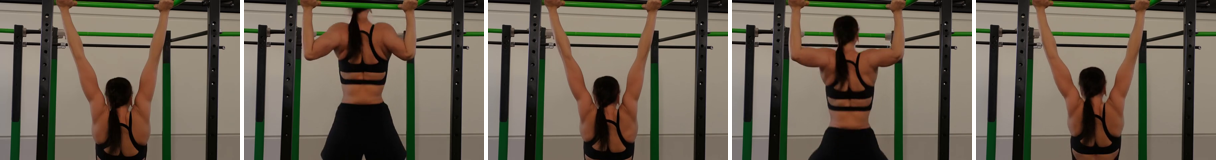}}
\put(11, 49){\rotatebox{90}{CogVideoX}}
\put(2, -4){\rotatebox{90}{CogVideoX}}
\put(16, 0){\rotatebox{90}{+\textbf{TeDiO}}}
\put(375,45){\fbox{\parbox{114pt}{
                A woman engaging in a challenging workout routine, performing pull-ups on green bars.
            }}}
\end{picture}
\end{minipage}
    \caption{Across \textbf{VideoJAM-Bench}~\cite{chefer2025videojam} prompts, \textbf{TeDiO} produces more temporally stable and smoother videos than the base models.}
    \label{fig:videojam}
\end{figure*}

Quantitative results on VBench VideoJAM-Bench are reported in Table~\ref{tab:vbench} and Table~\ref{tab:videojam}, and qualitative examples from VideoJam-Bench~\cite{chefer2025videojam} are illustrated in Fig.~\ref{fig:videojam}.



Across both models (Wan2.1 and CogVideoX) and both benchmarks, TeDiO consistently:

\begin{itemize}
\item \textbf{Improves motion Smoothness}.
\item \textbf{Reduces Temporal Flickering}.
\item \textbf{Increases Subject \& Background Consistency}.
\item \textbf{Achieves Better Aesthetic and Image Quality of Frames}
\end{itemize}

\begin{figure}[h]
\centering
\begin{minipage}{\linewidth}
\begin{picture}(60,210)

\put(0,85){\includegraphics[width=\linewidth,trim={0 0cm 0 0},clip]{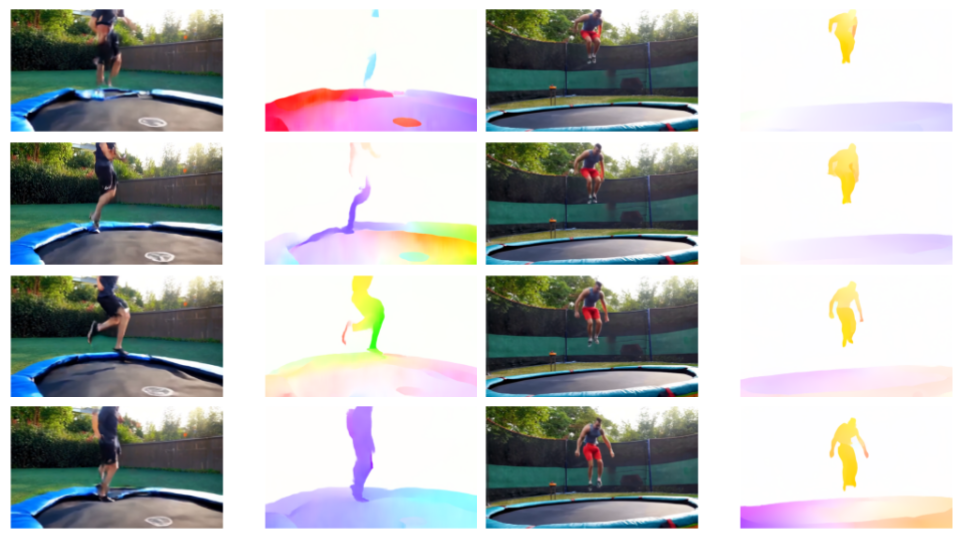}}
\put(25,0){\includegraphics[width=0.77\linewidth,trim={0 3.5cm 0 0},clip]{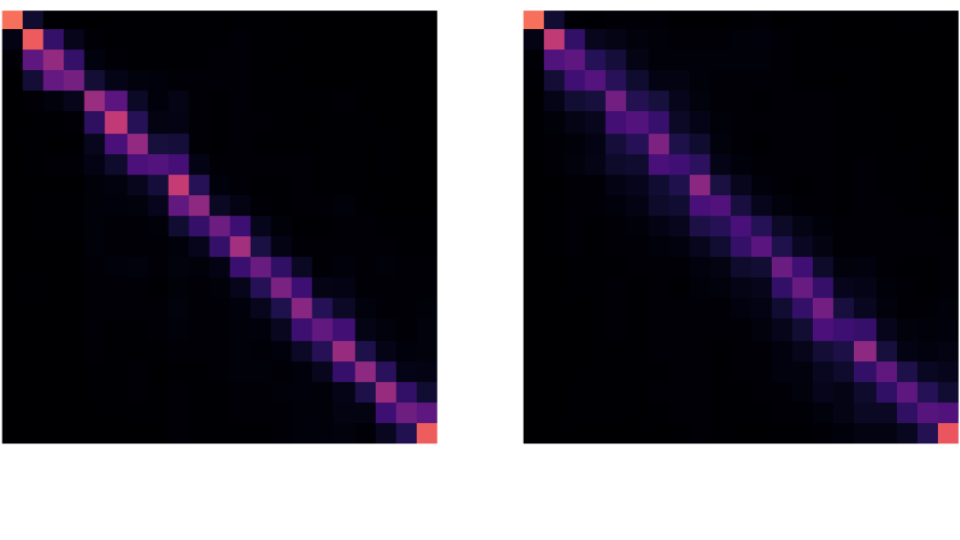}}
\put(17, 17){\rotatebox{90}{Wan2.1}}
\put(212, 65){\rotatebox{270}{Wan2.1+\textbf{TeDiO}}}
\end{picture}
\end{minipage}
\caption{\textbf{TeDiO} improves motion coherence, flow consistency, and smooths diagonal spikes, while preserving motion patterns.}
\label{fig:before_after}
\vspace{-0.5cm}
\end{figure}

These results demonstrate enhanced frame-to-frame stability and reduced representational drift - \textbf{crucially}, achieved without the need for model fine-tuning or additional supervision. As expected, Dynamic Degree (DD) decreases, reflecting the known trade-off between temporal smoothness and motion magnitude~\cite{huang2024vbench,chefer2025videojam,shaulov2025flowmo}. However, \textbf{higher DD does not necessarily imply better motion quality} - it can reflect jitter rather than intentional movement, as TeDiO \textbf{removes high-frequency temporal noise} that disproportionately inflates DD. Notably, real WebVid videos - whose motion profile closely resembles VBench prompts - achieve only $\mathbf{44.13}$ average DD~\cite{huang2024vbench} and TeDiO's VBench results lie comfortably within this range. Similarly, TeDiO's VideoJAM-Bench results remain close to the more dynamic UCF-101 regime (76.83~\cite{soomro2012ucf101}), enforcing coherence without oversmoothing.

Fig.~\ref{fig:before_after} further supports this: without TeDiO, optical flows oscillate rapidly between consecutive frames, with rapid color changes revealing unstable, jittery motion despite high DD; with TeDiO, flows are temporally consistent while preserving the overall motion structure. Correspondingly, temporal attention maps show diagonal motion patterns are retained while irregular high-magnitude spikes are suppressed.

\begin{table*}[]
    \centering
    \caption{\textbf{VideoJAM-Bench~\cite{chefer2025videojam} results.} TeDiO improves smoothness, consistency, and perceptual quality metrics across both backbones.}
    \begin{tabular}{l|cccccc}
    \toprule
         & \begin{tabular}{@{}c@{}}Motion \\ Smoothness \end{tabular} $\uparrow$ & \begin{tabular}{@{}c@{}}Dynamic \\ Degree \end{tabular} $\uparrow$ & \begin{tabular}{@{}c@{}}Subject \\ Consistency  \end{tabular} $\uparrow$ & \begin{tabular}{@{}c@{}}Background \\ Consistency \end{tabular} $\uparrow$ & \begin{tabular}{@{}c@{}}Aesthetic \\ Quality \end{tabular} $\uparrow$ & \begin{tabular}{@{}c@{}}Imaging \\ Quality \end{tabular} $\uparrow$ \\
         \midrule
         Wan2.1~\cite{wan2025}& 97.59 & \textbf{81.25} & 93.10 & 95.41 &  58.35 & 66.38  \\
          +\textbf{TeDiO (ours)} & \textbf{98.21} & 70.31 &  \textbf{94.78} & \textbf{96.24} & \textbf{59.22} & \textbf{67.84}\\
         \midrule
         CogVideoX~\cite{yang2024cogvideox}& 97.40 & \textbf{96.88} & 89.85 & 92.86 &  53.70 & 60.81\\
          +\textbf{TeDiO (ours)} & \textbf{98.06} & 89.84 &  \textbf{92.15} & \textbf{94.01} & \textbf{54.01} & \textbf{61.63}\\
         \bottomrule
    \end{tabular}
    \label{tab:videojam}
\end{table*}

\pgfplotstableread{
Label      Baseline  Tie  TeDiO
Wan2.1     26.47     18.18   55.35
CogVideoX  21.94     20.61   57.45
}\userstudy

\begin{figure}[t]
\centering
\begin{tikzpicture}
\begin{axis}[
    xbar stacked,
    xmin=0, xmax=100,
    xtick=\empty,
    width=0.95\linewidth,
    height=2.82cm,
    bar width=25pt,
    ytick=data,
    yticklabels from table={\userstudy}{Label},
    yticklabel style={yshift={ifthenelse(\ticknum==0,1ex, ifthenelse(\ticknum==1,-0.9ex, 0ex))                      
        }},
    legend style={
        at={(0.5,1.1)},
        anchor=south,
        legend columns=-1,
        draw=none
    },
    nodes near coords,
    point meta=explicit symbolic,
    nodes near coords align={left},      every node near coord/.append style={
        xshift={%
    -0.0*\pgfplotspointmeta
    \ifnum\pdfstrcmp{\pgfplotspointmeta}{21.94}=0
        -5pt
    \else
        \ifnum\pdfstrcmp{\pgfplotspointmeta}{26.47}=0
            -8pt
        \fi
    \fi
    \ifnum\pdfstrcmp{\pgfplotspointmeta}{20.61}=0
        -2pt
    \else
        \ifnum\pdfstrcmp{\pgfplotspointmeta}{18.18}=0
            -2pt
        \fi
    \fi
    \ifnum\pdfstrcmp{\pgfplotspointmeta}{55.35}=0
        -38pt
    \else
        \ifnum\pdfstrcmp{\pgfplotspointmeta}{57.45}=0
            -42pt
        \fi
    \fi
},
        yshift={ifthenelse(\coordindex==0,1ex, ifthenelse(\coordindex==1,-1ex, 0ex))                      
        },
        black
    },
]
    \addplot [fill=red!50] table [x=Baseline, y expr=\coordindex, meta=Baseline] {\userstudy};
    \addplot [fill=gray!40] table [x=Tie, y expr=\coordindex, meta=Tie] {\userstudy};
    \addplot [fill=green!50] table [x=TeDiO, y expr=\coordindex, meta=TeDiO] {\userstudy};
    \legend{Baseline, Tie, Baseline+\textbf{TeDiO}}
\end{axis}
\end{tikzpicture}
\caption{\textbf{User preference study on temporal coherence.} Participants selected which video had better temporal consistency. }

    \label{fig:user}
\end{figure}

\textbf{User Study.} Quantitative metrics offer important diagnostic signals but do not always reflect human perception of temporal smoothness. To complement automatic evaluation, we conducted a perceptual study using VideoJAM-Bench~\cite{chefer2025videojam}. $10$ participants on Prolific~\cite{prolific} viewed paired video generations and for each pair answered a single question: \emph{“Which video has better temporal coherence?”} As shown in Figure~\ref{fig:user}, our method is strongly preferred over both base models, with markedly higher selection rates.

\begin{table*}[t]
    \centering
    \caption{\textbf{Ablation studies} of different TeDiO hyperparameters on VideoJam-Bench~\cite{chefer2025videojam} prompts with Wan2.1-1.3B~\cite{wan2025}. MS=\textit{Motion Smoothness}, DD=\textit{Dynamic Degree}, SC=\textit{Subject Consistency}, BC=\textit{Background Consistency}}
    \label{tab:three_tables}
    \begin{subtable}[t]{0.31\textwidth} 
        \centering
        \caption{Timestep ablations}
        \label{tab:table1}
        \begin{tabular}{@{}c|cccc@{}}
            \cmidrule[\heavyrulewidth]{2-5}
             & MS & DD & SC& BC \\
            \midrule
            5 &  98.17 & \textbf{74.21} & 94.50 & 96.17\\
            12 &  \textbf{98.21} & 70.31 & \textbf{94.78} & \textbf{96.24}\\
            21 &  \textbf{98.21} & 71.88 & 94.76 & 96.13\\
            33 &  \textbf{98.21} & 71.09 & 94.65 & 96.13\\
            50 &  \textbf{98.21} & 69.53 & \textbf{94.78} & 96.22\\
            \bottomrule
        \end{tabular}
    \end{subtable}
    \hfill 
    \begin{subtable}[t]{0.34\textwidth} 
        \centering
        \caption{Patch number ablations}
        \label{tab:table2}
        \begin{tabular}{@{}c|cccc@{}}
            \cmidrule[\heavyrulewidth]{2-5}
            & MS & DD & SC& BC \\
            \midrule
            1 & 98.21 & \textbf{70.31} & 94.78 & 96.24 \\
            10 & 98.40 & 62.50 & 95.06 & 96.27 \\
            100 & 98.51 & 53.13 & 95.60 & 96.68 \\
            1000 & 98.58 & 48.44 & 95.90 & 96.81  \\
            18720 & \textbf{98.88} & 37.50 & \textbf{96.10} & \textbf{97.03} \\
            \bottomrule
        \end{tabular}
    \end{subtable}
    \hfill
    \begin{subtable}[t]{0.30\textwidth} 
        \centering
        \caption{Iteration number ablations}
        \label{tab:table3}
        \begin{tabular}{@{}c|cccc@{}}
            \cmidrule[\heavyrulewidth]{2-5}
            & MS & DD & SC& BC \\
            \midrule
            1 & 97.81 & \textbf{79.69} & 93.78 & 95.56\\
            2 & 98.07 & 76.56 & 94.36 & 95.96\\
            3 & 98.21 & 70.31 & 94.78 & 96.24\\
            4 & 98.31 & 64.84 & 94.93 & 96.24\\
            5 & \textbf{98.44} & 65.63 & \textbf{95.08} & \textbf{96.35}\\
            \bottomrule
        \end{tabular}
    \end{subtable}
\end{table*}

\subsection{Ablation Studies}

To further analyze TeDiO’s design choices and their impact on performance, we conduct ablation studies on several hyperparameters: (i) the number of optimized timesteps ($t$), (ii) the number of top-$k$ jerky patches selected for optimization, and (iii) the number of TeDiO update iterations. Table~\ref{tab:three_tables} summarizes the results. We additionally analyze which diffusion transformer block contributes most to temporal coherence (see supplementary material). 

Optimizing early timesteps (around $t{=}12$) yields the largest improvement in temporal smoothness, as optimizing later steps provides diminishing returns. Selecting a small number of jerky patches (e.g., $k{=}1$) offers a good trade-off between motion refinement and spatial diversity. Finally, applying TeDiO for three iterations is sufficient—further iterations bring only marginal improvement while slightly increasing runtime.

\subsection{Comparison with inference-time latent optimizations to improve coherence}

The most related approach to ours is \textbf{FlowMo}~\cite{shaulov2025flowmo}, which performs inference-time optimization of the input latent by minimizing the patch-wise variance of DiT outputs across frames.
While effective in reducing local inconsistencies, FlowMo requires costly gradient backpropagation through the full DiT model, making it computationally expensive and requires a minimum of 2 GPUs.

In contrast, \textbf{TeDiO} optimizes within a compact temporal subspace - guided by attention-derived motion cues - achieving higher temporal coherence and consistency with negligible runtime overhead. As shown in Table~\ref{tab:flowmo}, evaluated on VideoJAM-Bench~\cite{chefer2025videojam} prompts using Wan2.1-1.3B~\cite{wan2025}, TeDiO yields consistently larger gains ($\Delta$) in motion smoothness ($\Delta_{MS}$), subject consistency ($\Delta_{SC}$), and background consistency ($\Delta_{BC}$), while incurring only a $\times1.1$ runtime overhead over the base model and running comfortably on a single GPU. This highlights that TeDiO achieves superior temporal refinement without the heavy computational cost typical of gradient-based latent optimization methods.

\begin{table}[t]
    \centering
    \caption{\textbf{Comparison with FlowMo~\cite{shaulov2025flowmo}} on VideoJAM-Bench prompts. $\Delta$ shows improvement over original base model}
    \begin{tabular}{c|cc}
    \toprule
    & \multicolumn{2}{c}{Wan2.1-1.3B \cite{wan2025}}\\

         &  +FlowMo\cite{shaulov2025flowmo} & +\textbf{TeDiO (ours)}\\
         \midrule
         $\Delta_{MS}\uparrow$ & \textcolor{Green}{+0.06} & \textcolor{Green}{+\textbf{0.62}}\\
         \midrule
         $\Delta_{SC}\uparrow$ & \textcolor{Red}{-0.69} & \textcolor{Green}{+\textbf{1.68}} \\
         \midrule
         $\Delta_{BC}\uparrow$ & \textcolor{Red}{-0.34} & \textcolor{Green}{+\textbf{0.83}} \\
         \midrule
         \begin{tabular}{@{}c@{}} Runtime \\overhead \end{tabular} $\downarrow$ &  $\times2.39$& $\times\textbf{1.1}$\\
         \midrule
         \begin{tabular}{@{}c@{}} Minimum \\GPUs needed \end{tabular} $\downarrow$ &  2& \textbf{1}\\
         \bottomrule
    \end{tabular}
    \label{tab:flowmo}
\end{table}

\section{Conclusion}
We presented \textbf{TeDiO}, a training-free temporal optimization framework that improves video diffusion models purely at inference time. By leveraging temporal patterns implicitly captured in attention maps, TeDiO refines motion smoothness and perceptual consistency without retraining or architectural modification. Across Wan2.1 and CogVideoX, our approach delivers substantial gains in temporal coherence, maintaining image quality and runtime efficiency. This demonstrates that improving video realism need not rely on scaling model capacity, but rather on exploiting temporal structure already present in existing diffusion transformers.

\textbf{Limitations.}
While TeDiO effectively enhances temporal consistency, it does not address semantic correctness or prompt alignment errors inherent in base models. Its benefit also diminishes for extremely long videos or those requiring scene transitions. TeDiO is further less effective for intrinsically difficult motions requiring precise long-range coordination - such as acrobatic or rare articulated dynamics - as attention regularization alone cannot recover patterns poorly represented in the underlying model. Finally, TeDiO requires access to intermediate attention maps, which may not be exposed in certain closed-source implementations.

\newpage
{
    \small
    \bibliographystyle{ieeenat_fullname}
    \bibliography{main}
}


\end{document}